# REDBEE: A Visual-Inertial Drone System for Real-Time Moving Object Detection

Chong Huang, Peng Chen, Xin Yang, and Kwang-Ting (Tim) Cheng, *Fellow, IEEE*

*Abstract*— Aerial surveillance and monitoring demand both real-time and robust motion detection from a moving camera. Most existing techniques for drones involve sending a video data streams back to a ground station with a high-end desktop computer or server. These methods share one major drawback: data transmission is subjected to considerable delay and possible corruption. Onboard computation can not only overcome the data corruption problem but also increase the range of motion. Unfortunately, due to limited weight-bearing capacity, equipping drones with computing hardware of high processing capability is not feasible. Therefore, developing a motion detection system with real-time performance and high accuracy for drones with limited computing power is highly desirable.

In this paper, we propose a visual-inertial drone system for real-time motion detection, namely REDBEE, that helps overcome challenges in shooting scenes with strong parallax and dynamic background. REDBEE, which can run on the state-of-the-art commercial low-power application processor (e.g. Snapdragon Flight board used for our prototype drone), achieves real-time performance with high detection accuracy. The REDBEE system overcomes obstacles in shooting scenes with strong parallax through an inertial-aided dual-plane homography estimation; it solves the issues in shooting scenes with dynamic background by distinguishing the moving targets through a probabilistic model based on spatial, temporal, and entropy consistency. The experiments are presented which demonstrate that our system obtains greater accuracy when detecting moving targets in outdoor environments than the state-of-the-art real-time onboard detection systems.

## I. INTRODUCTION

Camera drones have dramatically improved the convenience and cost effectiveness for a number of applications, such as obtaining an overview of a large gathering and protests, or tracking a moving target from above. Several laser-based detection systems have been employed in drones for these tasks [25, 26]; however, the equipment required for these systems is both heavy and energy-inefficient. Therefore, vision-based motion detection is more feasible for low-cost and energy-constrained camera drones and thus attracts great interest from researchers and developers.

Chong Huang is with Department of Electrical and Computer Engineering, University of California, Santa Barbara, Santa Barbara, CA 93106 USA (chonghuang@umail.ucsb.edu).
Peng Chen is with the College of Information and Engineering, Zhejiang University of Technology, Hangzhou 310023 China (chenpeng@zjut.edu.cn)
Xin Yang is with the Electronics and Information Engineering Department, Huazhong University of Science and Technology, Wuhan, Hubei 430074 China, (xinyang2014@hust.edu.cn).
Kwang-Ting (Tim) Cheng is with Hong Kong University of Science and Technology, Clear Water Bay, Kowloon, Hong Kong (timcheng@usk.hk).

Most of the real-time applications [20] on camera drones involve sending a video data stream back to a ground station with a high-end computing server. However, a fast and reliable data connection between the drone and the ground station is essential. Delayed and corrupted data is likely to affect the surveillance performance. The possibility of interception and interference limits the operational range of the drone. The availability of light-weight energy-efficient hardware (i.e. the latest ARM-based application processor) makes it possible for onboard computation with high performance. For example, Qualcomm Snapdragon Flight, a highly integrated board embedded with Snapdragon 801 processor, allows developers to support a broad range of applications for drones ranging from aerial photography to video surveillance.

TABLE I

RUNNING SPEED ON DIFFERENT PROCESSORS ON VIDEO STREAM 640x480 (FPS)

|  | Intel CORE i7-5600U 2.6GHz | ARM Snapdragon 801 2.4GHz |
|---|---|---|
| RV[17] | 4.54 | 1.12 |
| LRR[18] | 8.28 | 2.97 |
| SSC[19] | 0.42 | 0.10 |

However, many vision algorithms [17, 18, 19] cannot be directly ported to a low-power processor because of the processor's limited processing capability. For example, Table 1 shows that there exists a large gap in runtime performance for running some representative algorithms on Intel and ARM processors. These detection algorithms with high accuracy cannot achieve real-time performance (above 12fps) running in an embedded computing system based on ARM-based application processors.

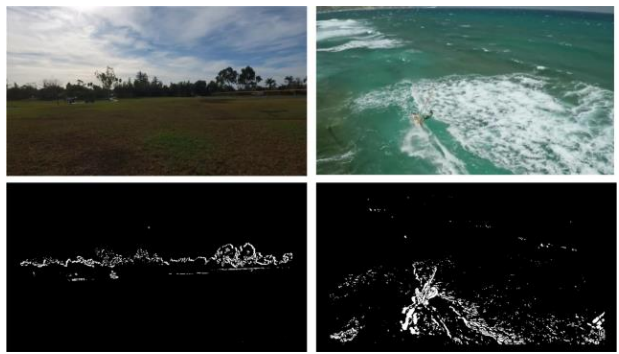

Fig. 1. The problem tackled in this paper is to produce pixel-wise segmentations that distinguish the moving objects from parallax (left column) and dynamic background (right column). The top row shows raw video from a moving camera drone. The binary masks in the bottom row are the motion detection results from detection system AURORA [22].

Contrary to the above sophisticated approaches, some researchers developed light-weight approaches (e.g. background subtraction) as baseline to build real-time surveillance system on an embedded system [20, 21, 22]. These approaches are based on the assumption that the background can be approximated by a plane. However, the detected results are often polluted by the noise caused by parallax and dynamic background. As shown in Fig. 1(left), the trees on the background are misidentified as moving objects while the drone is landing vertically. Fig. 1(right) shows that the resulting detection region contains large background of rippling waves. Therefore, the key challenge is how to overcome such problems and achieve high accuracy for drones with limited resources for computing.

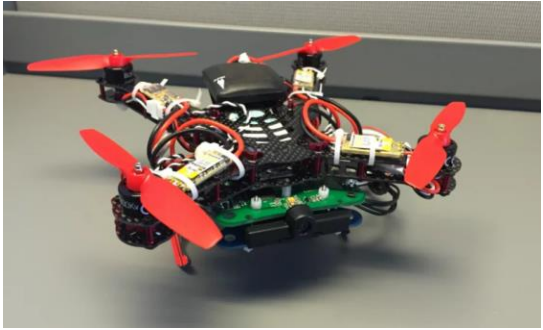

Fig. 2. Our prototype drone based on Qualcomm Snapdragon Flight

In this paper, we propose a real-time motion detection system REDBEE and demonstrate its performance using a prototype drone based on the Qualcomm Snapdragon Flight platform (see Fig.2). We demonstrate that REDBEE can perform real-time and robust onboard motion detection for scenes with strong parallax and dynamic background. REDBEE includes two new techniques: First, the background in an aerial video is modeled by two layers: a ground plane and a front plane as shown in Fig. 3. This dual-plane model can handle the parallax without losing detection accuracy because the depth disparity on each plane is relatively low compared with the distance from the camera drone. Second, a novel probabilistic model based on spatial, temporal and entropy properties is proposed to describe the uncertainty in foreground appearance, which can distinguish the independently moving objects from dynamic background.

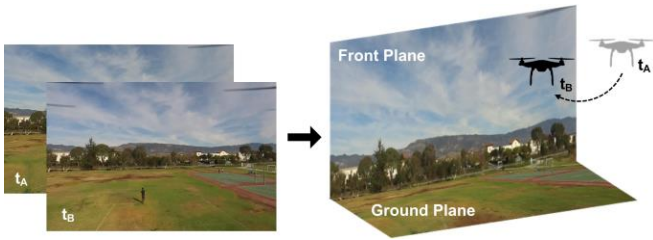

Fig. 3. Dual-plane background modeling

Experimental results demonstrate that our system can accurately detect moving targets in different scenarios, even with strong parallax effects and with backgrounds which are highly dynamic.

The rest of the paper is organized as follows: Sec. 2 reviews the related work. Sec. 3 presents details of our system. In Sec. 4, we report experimental results and Sec. 5 concludes the paper.

## II. RELATED WORK

Several approaches have been proposed for video surveillance on drone platforms. Some tracking system is based on the object detection with prior knowledge of the target. Sadlier et al [23] provides continuous robust vehicles tracking throughout the scenes of aerial video footage. Susan et al [24] presents an approach for the human detection and tracking using multi-sensor modalities including 3D Lidar and long-wave infrared video. These detection systems can achieve high accuracy in some specified applications (e.g. traffic control), but they are not robust for various target appearance and pose. The data-dependent training limits the deployment in a wider range of environments.

The alternate video surveillance systems involve motion detection based on the analysis of continuous frames. If the drone performs the detection during flight, the camera motion compensation is necessary for the background modeling. Michael et al [20] developed the video exploitation drone system ABUL for detection and classification of moving objects. Their system relies on fast and reliable video data transmission between the drone and the ground station. In some cases, delayed and corrupted data is likely to affect the surveillance performance. The possibility of interception and interference limits the operational range of the drone.

Development of low-power processor and light-weight motion detection algorithms make it possible to achieve real-time onboard computation. Gonzalo et al [21] developed a system to detect and track moving objects from camera drone embedded with Intel Atom processor (1.6 GHz). The intuition behind this system is to group the inconsistent optical flow with camera motion as moving objects. However, this method is sensitive to the dynamic background (e.g. waves). Plamen et al [22] designed and implemented a motion detection system AURORA on ADLINK 850 single board computer with Intel Core 2 Duo Processor (2.26GHz) mounted on a DJI hexacopter S800. Recursive density estimation (RDE) proposed in their system can efficiently remove the noise caused by randomly changing background. However, they assume that the background is approximated as a plane, and all of the motion detection is based on one homography matrix corresponding to this plane. As a result, the noise caused from parallax cannot be removed.

Methods mentioned above are primarily based on vision techniques. Few studies exist for inertial-aided motion detection. In [9], the authors proposed a hybrid vision-inertial system to detect the moving rigid objects from the video taken by an optical flow camera looking forward to the planar ground. This method shares the similar idea with the algorithm of [21]. In addition, they improve the detection performance by adding inertial data for motion compensation. However, this method is not suitable for complex, large-scale outdoor environments which have strong parallax effects and dynamic backgrounds.

## III. SYSTEM

In this section, we introduce our video surveillance system. In order to overcome the challenges (i.e. strong parallax and dynamic background) in surveillance scenarios, we propose a light-weight robust motion detection method which can run efficiently in an embedded computing platform such as Qualcomm Snapdragon Flight board. Our method is based on the assumption that background in an aerial video is modeled by two layers: a ground plane and a front plane as shown Fig. 3. This dual-plane model can handle the parallax without losing detection accuracy because the depth disparity on each plane is relatively low compared with the distance from camera drone. The proposed system includes two techniques: 1) An inertial-aided dual-plane homography estimation is proposed to address the strong parallax. 2) A probabilistic model based on spatial, temporal and entropy consistency is proposed to distinguish the moving targets from dynamic background.

Based on the dual-plane model, our framework is illustrated in Fig. 4. We provide details of both modules: inertial-aided moving object detection (Sec.3.1), and foreground refinement (Sec. 3.2).

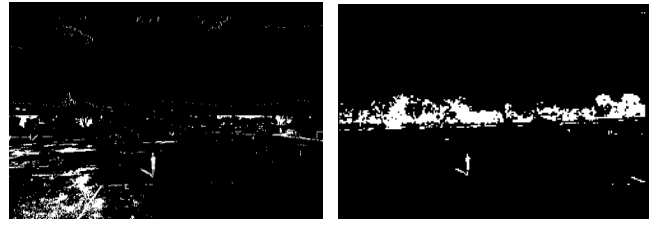

Fig. 5. The moving objects result on the image in Fig.3 (white region represents the identified moving objects) from front homography (left) and ground homography (right).

However, this vision-based multi-layer homography estimation becomes ineffective as the percentage of outliers from RANSAC is high, resulting in the corruption of homography estimation, especially homography based on the ground plane. In addition, it often fails to achieve satisfactory point tracking performance, especially for objects with large motion blurs caused by fast movement and long exposure time of the camera [12].

To address the limitation of the vision-based results, we propose to use a priori knowledge of the scenario and the inertial data obtained from IMU to calculate the homography based on the ground plane. IMUs can measure the angular and translational information of a camera with few restrictions on the object appearance. Specifically, consider a drone flying on a flat plane (see Fig. 6) with normal vector $n = (0, 0, 1)^T$. The camera on the drone looks forward to the plane at time $t_A$ with the initial rotation $R_A$ and height $h_A$, with respect to the IMU coordinate system $F_{IMU}$, then a rotation $R$ and translation $T$ occur at time $t_B$ with respect to the camera coordinate system $F_A$. Given the camera calibration matrix $K$, the homography matrix between $t_A$ and $t_B$ can be expressed in the following equation [13]:

$$H_I = KR_A(R + \frac{1}{h_A} nT)K^{-1} \qquad (1)$$

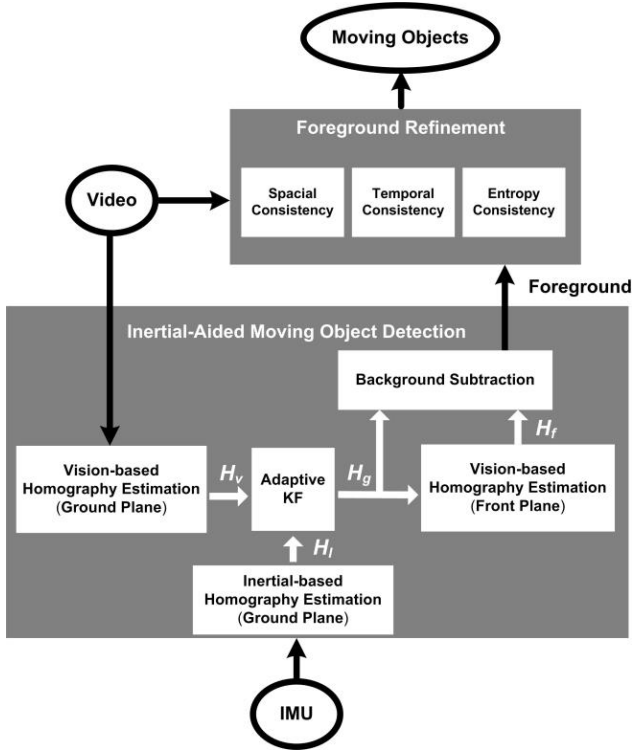

Fig. 4. The pipeline of our proposed method

### A. Inertial-Aided Moving Object Detection

We use a light-weight multi-layer homography estimation [2] to detect the foreground. First, a subset of the putative correspondences is computed by RANSAC [16] to fit one specific homography. Second, after excluding the previously calculated subset, we perform RANSAC again to find the next layer of homography for the residual feature points. Third, we use the homography matrix to warp the current frame into the previous one to compensate for the moving camera, and we run the background subtraction based on adaptive Gaussian Mixture Model [3] to get moving objects mask (see Fig. 5). Finally, we set the intersection of the results from both homography matrices as foreground. Assuming that an aerial image is dominated by the ground plane, the first homography is induced by the corresponding points on the ground plane and the second homography is based on front plane.

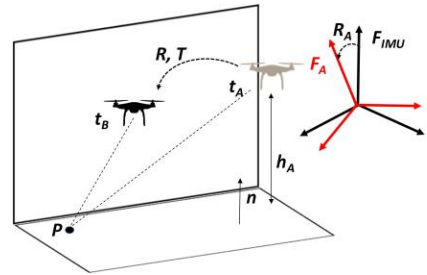

Fig. 6. Inertial ground plane homography estimation

The inertial ground plane homography estimation can be done very efficiently. However, the accuracy of inertial data usually degrades over time due to accumulated drifts caused by noises and bias in sensors. To combine the complementary advantages of vision and inertial information, we fuse visual and inertial homography in a Kalman Filter framework [14]. More specifically, we define the filter's state vector $x_t$ as the predicted homography from the initial time to time $t$. It is noted that $x_t$ has been formulated as a column vector with 8 degrees of freedom. The state-transition matrix $F_t$ is inertial homography from time $t$-1 and time $t$. The process model is essentially the multiplication of two homography for adjacent time periods.

$$x_t = F_t x_{t-1} + w_t \qquad (2)$$

$$x_t = H^{0:t} \qquad (3)$$

$$F_t = H_I^{t-1:t} \qquad (4)$$

where $w_k$ is a Gaussian distribution (i.e. $w_t \sim N(0, Q)$) modeling the inertial noise. In turn, the observation vector $z_t$ takes the values of the vision ground homography matrix from initial time to time $t$. The observation model is given by a vision-only model with an identity matrix $B_t$.

$$z_t = B_t x_t + v_t \qquad (5)$$

$$B_t = I_{8 \times 8} \qquad (6)$$

where $v_t$ depends on the point correspondence. Considering that the accuracy of vision homography estimation is related to the ratio of outliers to inliers, we use an adaptive Gaussian distribution (i.e. $v_t \sim N(0, P_t)$) to tune the visual noise parameters accordingly based on the ratio of outliers to inliers as Eq. 7. **#outliers(inliers)** refers to the number of outliers(inliers).

$$P_t = P_0 \left(1 + \frac{\# \, outliers}{\# \, inliers}\right) \qquad (7)$$

The homography update rule is similar to that of the standard Kalman filter. The difference is that the Gaussian variance $P_t$ of vision noise is adjusted at each iteration. More specifically, adjust the Gaussian variance $P_t$ of the observation noise proportional to the ratio of number of outliers to inliers. The Kalman gain is adjusted adaptively to de-emphasize the unreliable visual results, which will in turn improve the robustness of the hybrid system.

After the ground homography is obtained, we perform RANSAC again to find the secondary-plane homography for the outliers.

### B. Foreground Refinement

In many cases, the detected result is noisy. The conventional denoising method [15] cannot remove it efficiently. Specifically, it is hard to distinguish the dynamic background from other moving objects. In order to address this challenge, we propose a probability model to refine the foreground.

Our method is inspired by the observation that the object movement is smooth and coherent in both temporal and spatial domains [6]. That is to use the spatial and temporal properties of each pixel to refine the results. However, this pixel-wise method only eliminates the twinkle and discrete pixels with label "foreground", and it cannot handle the dynamic background. Here, we will add a grid-wise constraint to guide the foreground refinement.

The motion detection result (Fig. 7 (left)) computed by [3] shows that water waves are misidentified as moving objects. Fig. 7 (right) shows that the optical flow on sailboat is much more consistent compared with water waves. Based on information theory, the phenomenon happens because entropy of dynamic background is greater.

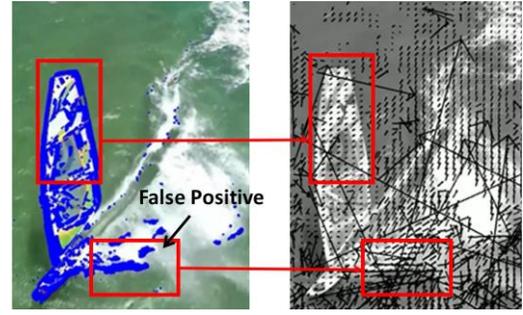

Fig. 7. (Left) motion detection result [24]. (Right) Optical flow

Based on the observation above, we can compute the entropy in a grid-based manner. More specifically, we divide each frame into multiple equal-sized grids, each of which covers 4×4 sample points. Because we perform KLT on the pixels uniformly and densely for optical flow estimation, an orientation histogram with 12 bins is formed for each grid, with each bin covering 30 degrees. As shown in Fig. 8, the orientation histogram of dynamic background (purple box) is smoother than that of the sailboat (red box), so we can use reciprocal entropy to measure the consistency of optical flow for the $k$th grid as:

$$E_t(k) = \frac{1}{\sum_{i=1}^{12} - P_b(i) \log P_b(i)} \qquad (12)$$

where $P_b(i)$ is the probability of the $i^{th}$ bin. The pixels within each grid share the same reciprocal entropy.

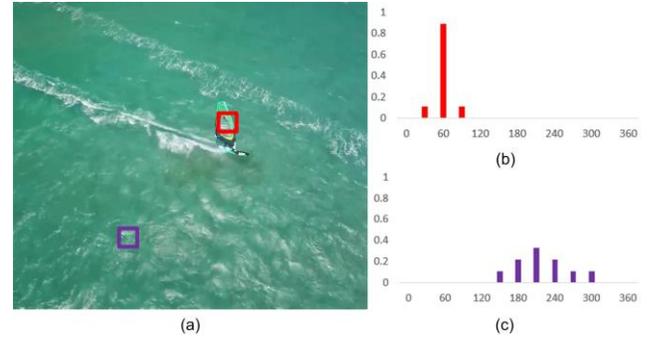

Fig. 8. (a) Video and two sampling boxes, which cover 4×4 sample points. (b) The quantized orientation histogram of optical flow within the red box and (c) the purple box.

We combine the temporal, spatial and entropy properties to detect moving objects.

Temporal property $P^T$ is defined as a recent history of the foreground at each pixel position as

$$P_t^T(n) = (1 - \alpha) P_{t-1}^T(n) + \alpha D_t(n) \qquad (13)$$

where $D_t$ is defined as a binary map in which $D_t(n) = 1$ if pixel $n$ belongs to the moving object at time $t$. $\alpha$ is the temporal learning rate.

Spatial property $P^S$ measures the coherency of nearby pixels of foreground as

$$P_t^S(n) = (1 - \beta)P_{t-1}^S(n) + \beta \frac{1}{w^2} \sum_{i \in N(n)} D_t(i) \quad (14)$$

$N(n)$ denotes a spatial neighborhood around pixel $n$, and $w^2$ is the area of neighborhood. $\beta$ is the spatial learning rate.

Similarly, $\gamma$ is the entropy learning rate entropy and property $P^E$ is written as:

$$P_t^E(n) = (1 - \gamma)P_{t-1}^E(n) + \gamma E_t(n) \quad (15)$$

The foreground probability $P_{FG}(n)$ is defined as multiplication of temporal, spatial and entropy properties, i.e.,

$$P_{FG}(n) = P_t^T(n) \times P_t^S(n) \times P_t^E(n) \quad (16)$$

The pixels with the foreground probability $P_{FG}(n)$ smaller than a threshold are removed from the initial results.

The silhouette of the moving object is further refined by standard morphological operations, including opening and closing operations [15].

## IV. EXPERIMENTS

In this section, we describe the dataset, the experimental setup and the measurement metrics used for our evaluation, followed by experimental results.

### A. Dataset

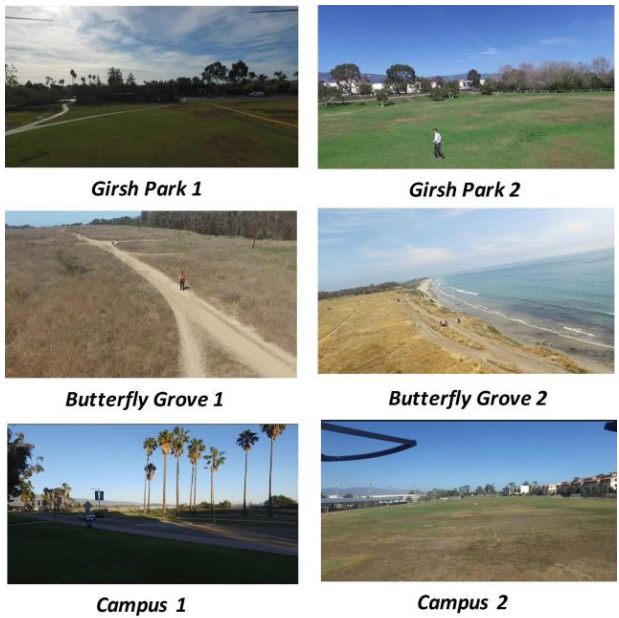

Fig. 9. The video clip samples.

We evaluated various methods using our own *Drone* dataset. The *Drone* dataset consists of 6 video clips including different scenarios and motions. All the clips are recorded in daytime in 24-bit RGB color (640 × 480) at 24 fps using a high resolution camera embedded on the drone. The length of each video is about 10 seconds. We manually annotated all visible moving objects in every five frame. Exemplar frames of video are shown in Fig. 9. For each video clip, we recorded the sensing data, synchronized with video frames, from the built-in IMU in the drone.

### B. Experimental Setup

We conduct the experiments on Snapdragon 801 (2.4GHz). The operating system is Linaro 4.8. In the experiments, we uniformly and densely sample points by every 16 pixels horizontally and vertically. We perform KLT feature tracker [11] on every sample point to calculate point correspondence. The 8×8 covariance matrix $Q$ is set as a diagonal matrix $diag(10^{-2}, 10^{-2}, 1, 10^{-2}, 10^{-2}, 1, 10^{-2}, 10^{-2})$, and the covariance matrix $P_0$ is set as the diagonal matrix $diag(10^{-3}, 10^{-3}, 10^{-4}, 10^{-3}, 10^{-3}, 10^{-3}, 10^{-3}, 10^{-3})$. In addition, the learning rate $\alpha$, $\beta$, $\Upsilon$ is set as 0.4, 0.4 and 0.6.

### C. Results

We evaluate the detection accuracy (Precision-Recall (PR) curve) in two sets of experiments: 1) Evaluation of different modules. 2) Comparison with onboard detection drone systems. For each experimental result, we sweep out the precision and recall curve for a video by classifying pixels as foreground or background with a varying foreground probability threshold. Given a certain threshold, the recall is defined as the number of the overlapping pixels between the detected region and groundtruth divided by the total number of the groundtruth. The precision is the number of the overlapping pixels between the detected results and groundtruth divided by the total number of the detected region. We evaluate the speed and accuracy by framerate and average error rate in 50% recall (Error@50%).

#### 1) Evaluation of different modules

We first evaluate the contribution of different modules for motion detection. In particular, we compare the proposed method with three simplified structures 1) without front plane 2) without inertial fusion 3) without foreground refinement.

TABLE II

EVALUATION OF DIFFERENT MODULES

| Method | Framerate (fps) | Error@50% |
|---|---|---|
| Proposed | 16.2 | **0.341** |
| w/o Front Plane | 18.7 | 0.437 |
| w/o Inertial Fusion | 17.1 | 0.378 |
| w/o Foreground Refinement | 22.9 | 0.381 |

Table 2 shows that dual-plane model, inertial fusion and foreground refinement can improve the performance without increasing too much computation. Fig. 10 gives the specific explanation. The system without front plane results in large false positive in the video clips with strong parallax (i.e. *Girsh Park* and *Campus 1&2*). The video clips *Butterfly Grove 1&2* have dynamic background (e.g. rippling bush and waves), foreground refinement can efficiently improve the performance. When the video clips (e.g. *Girsh Park 1&2*) have complex camera motion, inertial fusion can obviously achieve higher motion detection accuracy than vision-only system. Surprisingly, *Girsh Park 2* illustrates that foreground refinement slightly affects the detection results. The potential reason is that foreground refinement module is likely to filter the object with only several pixels.

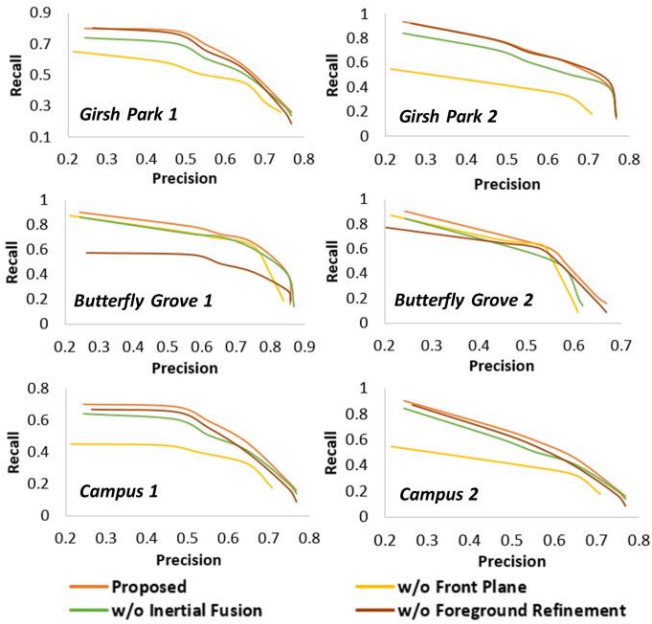

Fig. 10. The precision-recall curves of different structures

*2) Comparison with onboard detection drone systems*

In this section, we compare our proposed method with the state-of-the-art onboard detection drone systems: Meier et al [9] and AURORA [22]. We do not compare with the system [21] because the system [9] can be considered as the advanced version of the system [21] added with data fusion. We evaluate the detection system by speed, accuracy and memory usage.

TABLE III

COMPARING OUR METHOD WITH HYBRID VISION-INERTIAL METHOD

| Method | Framerate (fps) | Error@50% | Memory(MB) |
|---|---|---|---|
| Meier et al [9] | 20.6 | 0.442 | 8.5 |
| AURORA [22] | 19.1 | 0.422 | 9.1 |
| Proposed | 16.2 | **0.341** | 10.5 |

Table 3 and Fig. 11 show that the method of [9] and AURORA run slightly faster and consume less memory, but they generate more false positives (see Fig. 12). Both methods proposed in [9] and [22] approximate the background as a plane, so they are too oversimplified to model more complex background. Merier et al [9] clusters the consistent outlier groups in the optical flow of the image plane as the moving objects. This method does not need to model the background explicitly, but it cannot efficiently to remove the patch-wise noise caused by strong parallax. AURORA performs motion detection using RDE (Recursive Density Estimation), which is essentially a simplified background subtraction [3]. That is, once the background model is initialized based on background subtraction between the first and second frame, AURORA warps the background model into the new frames recursively which makes the foreground updated continuously. If there is the error of homography matrix calculation caused by dynamic or non-planar background, the accumulation errors will result in incorrect background modeling in the following frames. Fig.12 shows the qualitative results from different systems. We can see that our system can achieve high detection accuracy in the scenes with strong parallax and dynamic background.

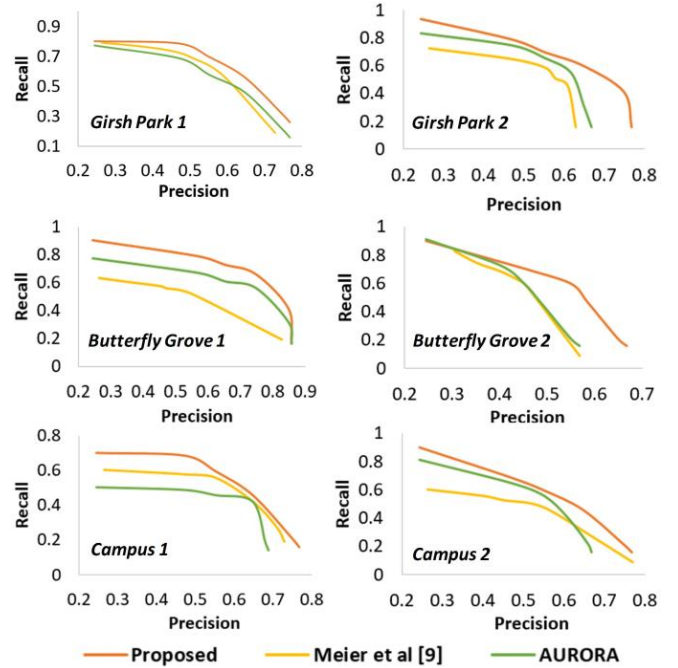

Fig. 11. The precision-recall curves of different methods

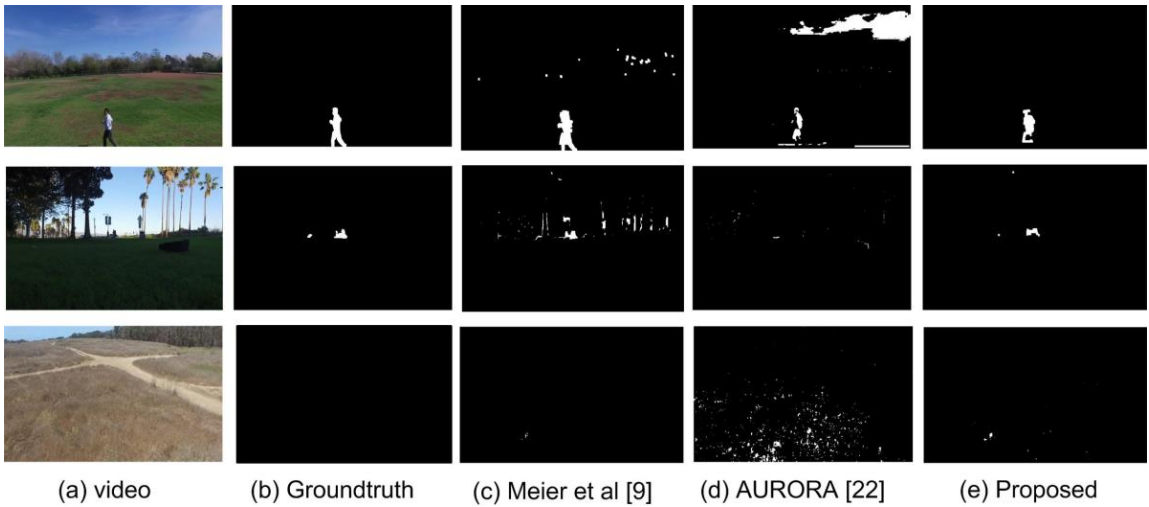

Fig. 12. Qualitative results on several images from different videos. From top to bottom: *Girsh Park 2*, *Campus 1* and *Butterfly Grove 1*.

## V. DISCUSSION

Model with ground plane and front plane: This seems to be suitable for outdoor scenarios with far-away horizon, but how will this behave indoors or within canyons or mountain areas? Does your two-stage homography estimation still work? Detection of moving rigid objects: What happens if the moving object stops? Is tracking (in a possible later step) then lost immediately, or can it be kept or recovered when beginning to move again.

It seems that front plane segmentation is beneficial if there is a front plane (horizon). If not (here: Butterfly Grove 1/2), there is no change. Same for the foreground refinement. So can you automatically detect when to use front plane and foreground optimization?

## VI. CONCLUSIONS

This paper presents an onboard moving objects detection system on camera drone. This system can effectively manage the tradeoff of the detection accuracy and computation speed on an energy-efficient hardware whose computing power is significantly inferior to desktop computers and servers. Real-time performance with high detection accuracy can be achieved due to the following proposed ideas: 1) We use dual-plane model to handle the strong parallax in the outdoor environment. The inertial information from drone platform is introduced to improve the homography estimation. 2) In order to distinguish the moving object from dynamic background, we develop a probabilistic model based on the observation in the spatial, temporal and entropy domains. Experimental results demonstrate that our system achieves higher detection accuracy while maintaining the similar framerate.